\pdfoutput=1
\documentclass[11pt,a4paper]{article}
\usepackage[hyperref]{acl2020}
\usepackage{times}
\usepackage{latexsym}

\usepackage{microtype}
\usepackage{latexsym} 
\usepackage{graphicx}
\usepackage{wrapfig}
\usepackage{amssymb}
\usepackage{color,xcolor,colortbl}
\usepackage{enumitem}
\usepackage{soul}
\usepackage{amsmath}
\usepackage{url}
\usepackage{algorithm}
\usepackage{algorithmic}
\usepackage[font={small}]{caption}
\usepackage{booktabs}
\usepackage{bm,bbm}
\usepackage{mathtools}
\usepackage{array}
\usepackage{multirow}
\usepackage{subcaption}
\usepackage{pbox}
\usepackage{pifont}
\usepackage{arydshln}
\usepackage{dblfloatfix}
\usepackage{relsize}

\newcommand\Tstrut{\rule{0pt}{2.6ex}}         

\definecolor{darkblue}{rgb}{0.0, 0.0, 0.55}
\setul{0.5ex}{0.3ex} 
\setulcolor{blue} 
\usepackage[T1]{fontenc}
\usepackage[scaled=.8]{beramono}

\aclfinalcopy 

\title{Automatic Summarization of Open-Domain Podcast Episodes}

\author{Kaiqiang Song,$^\spadesuit$ Chen Li,$^\diamondsuit$ Xiaoyang Wang,$^\diamondsuit$ Dong Yu,$^\diamondsuit$ Fei Liu$^\spadesuit$\\[0.8em]
$^\spadesuit$Computer Science Department, University of Central Florida\\
$^\diamondsuit$Tencent AI Lab, Bellevue, WA\\[0.6em]
\texttt{kqsong@knights.ucf.edu} \;
\texttt{\{ailabchenli,shawnxywang,dyu\}@tencent.com}\\
\texttt{feiliu@cs.ucf.edu}
}

\date{}

\begin{document}
\maketitle

\begin{abstract}

We present implementation details of our abstractive summarizers that achieve competitive results on the Podcast Summarization task of TREC 2020.
A concise textual summary that captures important information is crucial for users to decide whether to listen to the podcast.
Prior work focuses primarily on learning contextualized representations. 
Instead, we investigate several less-studied aspects of neural abstractive summarization, including
(i) the importance of selecting important segments from transcripts to serve as input to the summarizer;
(ii) striking a balance between the amount and quality of training instances;
(iii) the appropriate summary length and start/end points.
We highlight the design considerations behind our system and offer key insights into the strengths and weaknesses of neural abstractive systems.
Our results suggest that identifying important segments from transcripts to use as input to an abstractive summarizer is advantageous for summarizing long documents.
Our best system achieves a quality rating of 1.559 judged by NIST evaluators—an absolute increase of 0.268 (+21\%) over the creator descriptions.

\end{abstract}

\section{Introduction}

\begin{table}[t]
\setlength{\tabcolsep}{2pt}
\renewcommand{\arraystretch}{1.1}
\centering
\begin{scriptsize}
\textsf{
\begin{tabular}{|l|}
\hline
\rowcolor{gray!15}
\multicolumn{1}{|c|}{\textbf{Segments of a Podcast Transcript }}\\
\hdashline\Tstrut
What's good? Everybody is of all trades here with the game \\
Illuminati. Hope you guys are having a great day. So far. If you guys \\
didn't get a chance to check out our last video. Be sure to check out \\
the link down in the description box below. We are back for another \\
Triple Threat Sports podcast. Now before we get into the Super Bowl \\
edition of Triple Threat Sports podcast. I got to introduce you guys \\
to my co-host first co-host. Say what up C ewan's what up next you \\
guys know him as UT X JG to dine, but he's also known as the ...\\[0.2em]
\hdashline\Tstrut
The LA Rams supporter and he's going to the Superbowl. Say what\\
up, GG don't it? Feel so good though. It feels so good. The lone\\
person the triple threat Sports podcast by team is going to the shelf \\
and people are mad and we gonna talk about it man. We got we \\
definitely going to talk about it for sure. So at the time of this \\
recording we've already gone through the Pro Bowl which was \\
yesterday. I'm sure some of you guys watched it and you know \\
whoopty whoopty Doo I've ...\\[0.2em]
\hdashline\Tstrut
Interest in the Pro Bowl after like I turned I 15 but AFC 126 to 7, but \\
we're going to talk about these NFL Conference Championship \\
games. You got to between the Rams and the Saints which whoo, \\
boy, there's a lot of controversy behind that one and then the Chiefs \\
and the Patriots before we get to the smoke and everything like that \\
because jg's been handing them out all this week. Let's go ahead \\
and start going into the NFL conference championships for the \\
Patriots and the Chiefs now, I'll go with you ... \\[0.5em]
\hline
\hline
\rowcolor{gray!15}
\multicolumn{1}{|c|}{\textbf{Creator Description}} \\
\hdashline\Tstrut
The Guys are back for another Triple Threat Sports Podcast! This \\
time UTXJGTHEDON is giving out all the smoke as his Los Angeles \\
Rams is heading to the Super Bowl to face the New England Patriots. \\
--- Support this podcast: https://anchor.fm/triplethreatsportspodcast\\[0.5em]
\hline
\hline
\rowcolor{gray!15}
\multicolumn{1}{|c|}{\textbf{Our Summary}} \\
\hdashline\Tstrut
In this episode of the Triple Threat Sports Podcast, JG and UTX \\
discuss the NFL Conference Championship games between the \\
Patriots and Chiefs and the Rams and Saints. They also discuss \\
the controversy between the Chiefs and Patriots and what they  \\
had to say about it. The guys also give their predictions for the \\
Super Bowl and what teams they think are going to win the game.\\[0.5em]
\hline
\end{tabular}
}
\end{scriptsize}
\vspace{-0.1in}
\caption{
A snippet of the podcast transcript, its original creator description and an abstract produced by our summarizer.
}
\label{tab:example}
\vspace{-0.2in}
\end{table}

Podcast is a promising new medium for reaching a broad audience.
A podcast series usually feature one or more recurring hosts engaged in a discussion about a particular topic.\footnote{\url{https://en.wikipedia.org/wiki/Podcast}}
New platforms developed by Spotify, Apple, Google and Pandora encompass a wide variety of topics,
ranging from talk shows to true crime and investigative journalism.
Our data are provided by Spotify~\cite{trec2020podcastnotebook}, containing 100,000 podcast episodes comprised of raw audio files, their transcripts and metadata.
The transcription is provided by Google Cloud Platform's Speech-to-Text API.\footnote{\url{https://cloud.google.com/speech-to-text}}

We seek to generate a concise textual summary for any podcast episode, which a user might read when deciding whether to listen to the podcast. 
An ideal summary is required to accurately convey all the most important attributes of the episode, such as topical content, genre and participants.
It is best for the summary to contain no redundant material that is not needed when deciding whether to listen.
In Table~\ref{tab:example}, we show a snippet of the podcast transcript and its creator description. 
The snippet contains 3 segments, each corresponds to 30 seconds of audio.

\begin{figure*}
\centering
\includegraphics[width=5in]{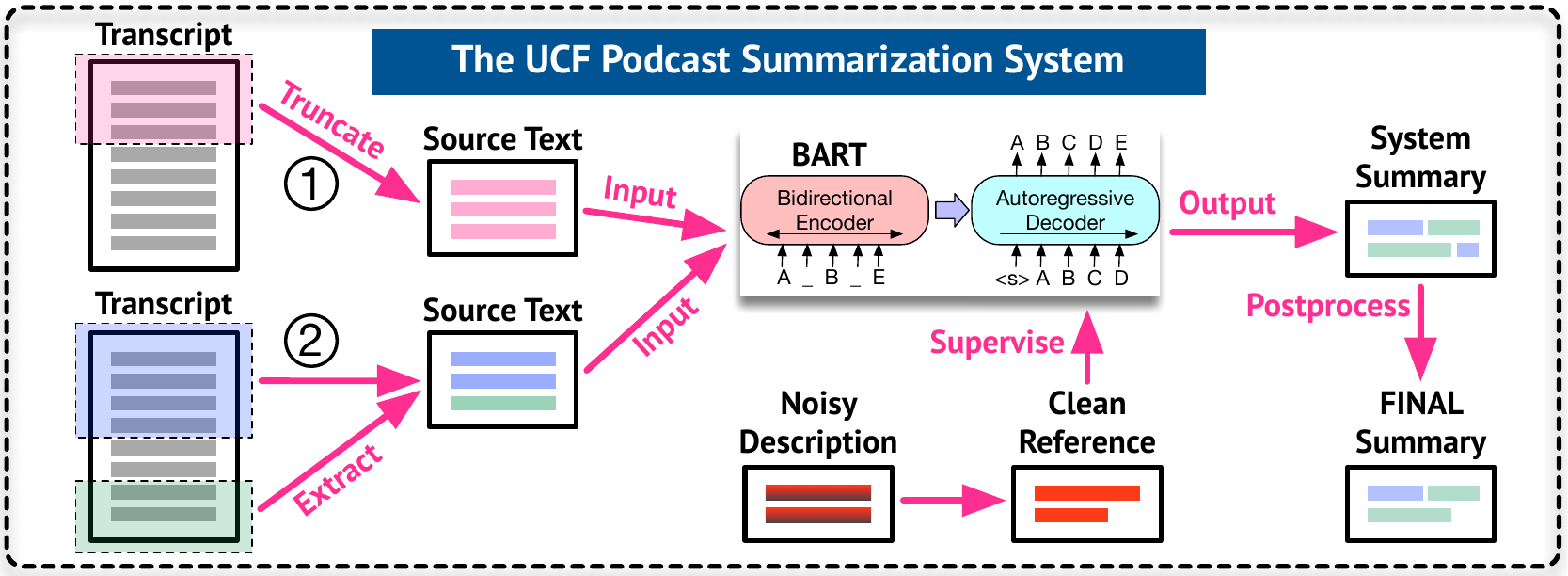}
\vspace{-0.05in}
\caption{
An illustration of our system architecture.
\textsf{\textbf{\scriptsize{UCF\_NLP1}}} truncates the transcript to a length of $L$=1,024 tokens,
before feeding it to the BART model to produce a concise abstract.
\textsf{\textbf{\scriptsize{UCF\_NLP2}}} produces an abstractive summary in a similar fashion. 
It enhances content selection by identifying important segments from the transcripts to serve as input to BART. The selected segments are limited to a length of $L$=1,024 tokens.
}
\label{fig:architecture}
\end{figure*}

The major challenge in performing podcast summarization includes
(a) the unique characteristics of spoken text.
Disfluencies and redundancies are abundant in spoken text; its information density is often low when compared to written text.
The podcasts are of various genres: monologue, interview, conversation, debate, documentary, etc.
and transcription is more challenging and noisier;
(b) the excessive length of transcripts.
It exceeds the limit imposed by many neural abstractive models.
A podcast transcript contains on average 80 segments and 5,743 tokens.
It serves as the input to a podcast summarization system to produce an abstract. 
The creator description is part of the metadata.
It contains 81 tokens on average and is used as the reference summary.

This work draws on our rich experience in summarizing meeting conversations~\cite{liu-etal-2009-unsupervised,liu-liu-2009-extractive,Liu:2013:IEEETrans,Koay:2020} and building neural abstractive systems~\cite{lebanoff-etal-2019-scoring,Lebanoff:2020:AACL,Song:2020:Copy}.
We have chosen an abstractive system over its extractive counterpart for this task, as neural abstractive systems have seen significant progress~\cite{Raffel:2019,lewis-etal-2020-bart,qi2020prophetnet}.
Not only can an abstract accurately convey the content of the podcast, but it is in a succinct form that is easy to read on a smartphone.
Our system seeks to fine-tune a neural abstractive summarizer with an encoder-decoder architecture~\cite{lewis-etal-2020-bart} on podcast data.
We especially emphasize content selection, where an extractive module is developed to select salient segments from the beginning and end of a transcript, which serve as the input to an abstractive summarizer.
In this work, we systematically investigate three crucial questions concerning abstractive summarization of podcast transcripts.
\begin{itemize}[topsep=5pt,itemsep=0pt]

\item Is it sufficient to feed the leading sentences of a transcript to the summarizer to produce an abstract, or are there advantages to be gained from selecting salient segments with an extractive module to use as input to the summarizer? (\emph{Content Selection})

\item Should we remove training pairs with noisy, low-quality reference summaries in entirety, or can we improve their quality, and thus strike a balance between the amount and quality of training examples? (\emph{The Quality of Reference})

\item What summary length would be most appropriate for podcast episodes to serve our goal of assisting users in deciding whether to listen to the podcast? (\emph{Summary Postprocessing})

\end{itemize}

\section{Our Method}
\label{sec:approach}

We aim to produce a concise textual summary from a podcast transcript that captures the most important information of the episode to help users decide whether to listen to that episode.
Our method makes use of podcast transcripts only but not raw audio.
It utilizes the BART model~\cite{lewis-etal-2020-bart} to condense a source text into an an abstractive summary,
which employs an encoder-decoder architecture. 
The model is pretrained using a denoising objective.
The source text is corrupted by replacing spans of text with mask symbols.
The encoder encodes the corrupted text using a bidirectional model, and the decoder learns to reconstruct the original source text by attending to hidden states of the final layer of the encoder using a cross-attention mechanism.

Our implementation is based on BART-\textsc{large}. 
The encoder and decoder each contain 12 layers of Transformer blocks.
The hidden state and embedding size is 1,024.
Byte-Pair Encoding (BPE; Sennrich et al., 2016\nocite{sennrich-etal-2016-neural}) is used to tokenize the source text. 
It has a vocabulary of 50,265 subword units. 
The BART model is fine-tuned on CNN (\textsf{\textbf{\footnotesize{bart-large-cnn}}}) then on podcast data;
the latter contain 79,262/500 examples for training and validation. 
Our system is evaluated on a test set with 1,027 examples.
Given a transcript, we compare two methods to generate the source text that serves as the input to BART, 
corresponding to two runs we submitted to the Podcast Challenge.
Our system architecture is illustrated in Figure~\ref{fig:architecture} and details are described as follows.

\begin{itemize}[topsep=5pt,itemsep=0pt]

\item \textsf{\textbf{\footnotesize{UCF\_NLP1}}} \quad
The transcript is truncated to a length of $L$=1,024 tokens.
The method takes the lead sentences of a transcript, feeds them to the BART model to produce a succinct abstractive summary that captures important information about the podcast episode.

\item \textsf{\textbf{\footnotesize{UCF\_NLP2}}} \quad
It produces an abstractive summary from a podcast transcript in a similar fashion. 
Crucially, the method enhances content selection by identifying summary-worthy segments from the transcript to serve as input to BART. The selected segments are limited to a length of $L$=1,024 tokens.

\end{itemize}

\subsection{Content Selection}
\label{sec:content}

We seek to empirically answer the question: ``\emph{Is it sufficient to feed the lead sentences of the transcript to an abstractive summarizer, or are there advantages to be gained from selecting salient segments with an extraction module to use as input to the summarizer?}''
We consider segments produced by the Google Speech-to-Text API as basic units of extraction, each corresponds to 30 seconds of audio.
We opt for segment- rather than sentence-based extraction for two reasons.
First, the information density of single utterances is often low.
In contrast, the segments are lengthier and more detailed.
They tend to have similar lengths and are less likely to be misclassified due to length variation.
Second, comparing to sentences, concatenating segments to form a source text that serves as the input to BART can help preserve the context of the utterances extracted from the transcript.

We introduce a hybrid representation for the $i$-th candidate segment that combines deep contextualized representations and surface features.
Particularly, each segment is encoded by RoBERTa~\cite{liu2019roberta}.
It contains 24 layers of Transformer blocks, has a hidden size of 1024 and 16 attention heads.
A special token \textsf{\scriptsize{[CLS]}} is added to the beginning of the segment and \textsf{\scriptsize{[SEP]}} is added to the end.
We use the output vector corresponding to the \textsf{\scriptsize{[CLS]}} token as the contextualized representation for the segment, denoted by $\bm{h}_i^{c} \in \mathbb{R}^D$.

A segment containing salient words is deemed to be important.
We measure word salience by its duration (in seconds) and TF-IDF score, which are orthogonal to contextualized representations and aim to capture the topical salience.
To characterize a segment using its containing words, we compute 12 feature scores for a candidate segment, 
including (a) the sum and average of word TF-IDF scores; (b) the sum and average of word durations;
(3) the average of word TF-IDF scores (and durations), limiting to 5/10/15/20 words per segment that yield the highest scores. 
Each feature score is discretized into a binary vector using a number of bins whose sizes are \{2, 3, 5, $\cdots$ 31, 37\} (12 prime numbers).
E.g., a feature score is mapped to a 2-dimensional  vector [0,1] (bin size=2) if its value is in the upper half of all values; otherwise it is [1,0].
By concatenating binary vectors of different bin sizes, and vectors corresponding to different feature scores, we obtain a 2,364-dimentional vector for each segment.
The vector is passed through a feedforward layer to generate a surface feature vector of size $D$, denoted by $\bm{h}_i^{s} \in \mathbb{R}^D$.

We take 33 segments from the beginning and 7 segments from the end of each transcript to be the candidate segments.
This amounts to a total of 40 segments per episode.
The selection is bounded by the GPU memory, but allows us to cover 81\% of the ground-truth summary segments. 
Each segment is characterized by its contextualized representations $\bm{h}_i^{c}$, surface features $\bm{h}_i^{s}$, and a position embedding $\bm{h}_i^{p}$, all of which are added up in an element-wise manner to serve as input to a 2-layer Transformer encoder, with a hidden size of $D$=1,024, 16 attention heads and no pretraining, to produce a vector for each candidate segment of the transcript.
Each vector is fed to a feedforward and a softmax layer to predict if the segment is salient.

The ground-truth segment labels are derived by comparing segments with creator descriptions.
We calculate the ROUGE-2 Recall score for a segment against any sentence of the creator description.
A segment is labelled as positive if the score is greater than a threshold ($\tau$=0.2), otherwise negative.
The positive-to-negative ratio is 1:18 among candidate segments, 
and no downsampling was performed.
Our preliminary results suggest that using a hybrid representation that combines surface features with contextualized representations for the segments can lead to an improvement in extraction performance (+0.53\% F-score).

\begin{table}[t]
\setlength{\tabcolsep}{2pt}
\renewcommand{\arraystretch}{1.15}
\centering
\begin{scriptsize}
\textsf{
\begin{tabular}{|l|}
\hline
\rowcolor{gray!15}
\multicolumn{1}{|c|}{\textbf{Creator Description}} \\
\hdashline\Tstrut
The Guys are back for another Triple Threat Sports Podcast! This \\
time UTXJGTHEDON is giving out all the smoke as his Los Angeles \\
Rams is heading to the Super Bowl to face the New England Patriots. \\
--- Support this podcast: https://anchor.fm/triplethreatsportspodcast\\[0.5em]
\hline
\hline
\rowcolor{gray!15}
\multicolumn{1}{|c|}{\textbf{Clean Reference Summary}} \\
\hdashline\Tstrut
The Guys are back for another Triple Threat Sports Podcast! This \\
time UTXJGTHEDON is giving out all the smoke as his Los Angeles \\
Rams is heading to the Super Bowl to face the New England Patriots. \\
\hline
\end{tabular}
}
\end{scriptsize}
\vspace{-0.05in}
\caption{
Our data cleansing method focuses on improving the quality of reference summaries using a number of heuristics, rather than eliminating noisy reference summaries in entirety.
It thus strikes a good balance between the quality and amount of training examples.
}
\label{tab:ref}
\vspace{-0.1in}
\end{table}

\begin{table*}
\setlength{\tabcolsep}{5pt}
\renewcommand{\arraystretch}{1.15}
\centering
\begin{scriptsize}
\textsf{
\begin{tabular}{|l|l|}
\hline\Tstrut
(\textbf{3}) \textbf{Excellent} & The summary accurately conveys all the most important attributes of the episode, which could include topical content, genre, \\
& and participants. In addition to giving an accurate representation of the content, it contains almost no redundant material \\
& which is not needed when deciding whether to listen. It is also coherent, comprehensible, and has no grammatical errors. \\
\hdashline
(\textbf{2}) \textbf{Good} & The summary conveys most of the most important attributes and gives the reader a reasonable sense of what the \\
& episode contains with little redundant material which is not needed when deciding whether to listen. Occasional \\
& grammatical or coherence errors are acceptable.\\
\hdashline
(\textbf{1}) \textbf{Fair} & The summary conveys some attributes of the content but gives the reader an imperfect or incomplete sense of what the \\
& episode contains. It may contain redundant material which is not needed when deciding whether to listen and may contain \\
& repetitions or broken sentences.\\
\hdashline
(\textbf{0}) \textbf{Bad} & The summary does not convey any of the most important content items of the episode or gives the reader an incorrect or \\
& incomprehensible sense of what the episode contains. It may contain a large amount of redundant information that is not \\
& needed when deciding whether to listen to the episode.\\[0.3em]
\hline
\end{tabular}
}
\end{scriptsize}
\vspace{-0.05in}
\caption{
The qualitative judgments performed by NIST.
The rating is a number from 0-3, with 0 being Bad and 3 being Excellent.
}
\label{tab:eval_criteria}
\end{table*}

\begin{table}
\setlength{\tabcolsep}{3pt}
\renewcommand{\arraystretch}{1.15}
\centering
\begin{scriptsize}
\textsf{
\begin{tabular}{|l|l|}
\hline
\textbf{Q1} & Does the summary include \textbf{names of the main people} (hosts, \\
& guests, characters) involved or mentioned in the podcast? \\[0.2em]
\hdashline
\textbf{Q2} & Does the summary give any \textbf{additional information} about \\
& the people mentioned (such as their job titles, biographies, \\
& personal background, etc)? \\[0.2em]
\hdashline
\textbf{Q3} & Does the summary include the \textbf{main topic(s)} of the podcast? \\[0.2em]
\hdashline
\textbf{Q4} & Does the summary tell you anything about \textbf{the format of} \\
& \textbf{the podcast}; e.g. whether it's an interview, whether it's a chat \\
& between friends, a monologue, etc? \\[0.2em]
\hdashline
\textbf{Q5} & Does the summary give you \textbf{more context on the title} \\
& of the podcast?\\[0.2em]
\hdashline
\textbf{Q6} & Does the summary contain \textbf{redundant information}? \\[0.2em]
\hdashline
\textbf{Q7} & Is the summary written in \textbf{good English}? \\[0.2em]
\hdashline
\textbf{Q8} & Are the \textbf{start and end of the summary} good sentence and \\
& paragraph start and end points? \\[0.2em]
\hline
\end{tabular}
}
\end{scriptsize}
\vspace{-0.05in}
\caption{
There are eight yes-or-no questions asked about the summary. 
The judgments are performed by NIST. 
An ideal summary should receive a ``yes'' (1) for all questions but Q6.
}
\label{tab:eval_questions}
\end{table}

\subsection{The Quality of Reference}
\label{sec:quality}

One of the significant challenges in abstractive summarization is the scarcity of labelled data.
While it is common practice to remove training examples containing noisy, low-quality reference summaries, it is not obvious whether this is the best path to take for data curation, as a significant amount of examples may be eliminated from the training set.
Thus, we raise the question: 
``\emph{Should we remove training examples with noisy low-quality reference summaries in entirety, or can we improve their quality, and thus strike a balance between the amount and quality of training examples?}''

The training data provided by the Podcast Challenge contain over 100,000 episodes and short descriptions written by their respective creators. 
The organizers find that about a third are less useful descriptions, and have since filtered out descriptions that are
(a) too long (greater than 750 characters) or short (less than 20 characters);
(b) too similar to other descriptions (cut-and-paste or template);
(c) too similar to its show description (no new info).
This practice results in 66,245 training examples, 
corresponding to a 34\% reduction of training data, which has a visible effect on performance.

Instead of eliminating noisy examples in entirety, we strive to enhance the quality of creator descriptions using heuristics.
Our goal is to identify sentences that contain improper content and remove them from the descriptions.
We compute a salience score for each sentence of the description by summing over word IDF scores.
A low IDF score indicates the word frequently appears in other episodes, and thus is uninformative.\footnote{
We perform data normalization by replacing URLs, Email addresses, @usernames, \#hashtags, digits and tokens that are excessively long (greater than 25 characters) with placeholders before computing word IDF scores. Only words occurring 5 times or more in the corpus and with IDF scores greater than 1.5 are considered when computing sentence salience scores.
}
We remove sentences if their salience scores are lower than a threshold ($\sigma$=10).
The remaining sentences of a creator description are concatenated to form a clean reference summary.
Our method results in 79,912 training examples.
It reduces the average length of the reference summary from 81 to 76 words.
In Table~\ref{tab:ref}, we show an example containing reference summaries before and after data cleansing.

\begin{table*}
\setlength{\tabcolsep}{6pt}
\renewcommand{\arraystretch}{1.15}
\centering

\begin{minipage}{\textwidth}
\begin{scriptsize}
\textsf{
\begin{tabular}{|l|r|rrrrrrrr|}
\hline
\rowcolor{gray!15}
& \textcolor{red}{\textbf{Quality}} & Q1: People & Q2: People & Q3: Main & Q4: Podcast & Q5: Title & Q6: Summ & Q7: Good & Q8: Start/End\\
\rowcolor{gray!15}
System & \textcolor{red}{\textbf{Rating}} & Names & Add Info & Topics & Format & Context & Redund & English & Points\\
\hline
\hline
DESC & 1.291 & 0.559 & 0.341 & 0.721 & 0.553 & 0.637 & 0.034 & 0.777 & 0.520\\
FILT & 1.307 & 0.581 & 0.380 & 0.704 & 0.531 & 0.603 & \textbf{0.028} & 0.782 & 0.609\\
\textcolor{red}{UCF\_NLP1} & 1.453 & 0.609 & 0.374 & \textbf{0.804} & \textbf{0.564} & \textbf{0.821} & 0.078 & 0.827 & \textbf{0.659}\\
\textcolor{red}{UCF\_NLP2} & \textcolor{red}{\textbf{1.559}} & \textbf{0.642} & \textbf{0.385} & 0.765 & \textbf{0.564} & 0.726 & 0.061 & \textbf{0.877} & 0.620\\
\hline
\end{tabular}
}
\end{scriptsize}
\caption{
The average results of human judgments for 179 testing summaries. The evaluation was performed by NIST assessors. 
An assessor quickly skimmed the episode, and made judgments for each summary of the episode.
``\textsf{\scriptsize{DESC}}'' represents the episode description.
``\textsf{\scriptsize{FILT}}'' is a ``filtered'' summary provided by Spotify. 
``\textsf{\scriptsize{UCF\_NLP}}$\star$'' are our system outputs.
Our best system ``\textsf{\scriptsize{UCF\_NLP2}}'' uses an additional extraction component to identify summary-worthy segments from the transcripts.
It achieves a quality rating of 1.559. 
This is an absolute increase of 0.268 (+21\%) over the episode description.
}
\label{tab:human_eval}
\end{minipage}
\vskip0.8\baselineskip
\begin{minipage}{\textwidth}
\begin{scriptsize}
\textsf{
\begin{tabular}{|l|r|rrrrrrrr|}
\hline
\rowcolor{gray!15}
& Quality & Q1: People & Q2: People & Q3: Main & Q4: Podcast & Q5: Title & Q6: Summ & Q7: Good & Q8: Start/End\\
\rowcolor{gray!15}
System & Rating & Names & Add Info & Topics & Format & Context & Redund & English & Points\\
\hline
\hline
DESC & 91.06 & 89.39 & 93.30 & 86.59 & \textbf{94.41} & 84.36 & 99.44 & 85.48 & 91.06\\
FILT & 89.39 & 92.18 & 92.74 & 84.92 & 92.74 & 82.12 & 99.44 & 86.03 & \textbf{96.65}\\
\textcolor{red}{UCF\_NLP1} & 93.30 & 97.21 & \textbf{97.21} & \textbf{93.86} & 93.86 & 91.06 & \textbf{100} & 93.30 & 93.30\\
\textcolor{red}{UCF\_NLP2} & \textcolor{red}{\textbf{96.65}} & \textbf{98.88} & 96.65 & 92.18 & 93.86 & \textbf{92.18} & \textbf{100} & \textbf{94.97} & 93.86\\
\hline
\end{tabular}
}
\end{scriptsize}
\caption{
Percentages of testing episodes (out of a total of 179) on which our system performs equal to or better than the majority baseline. 
The majority rating was obtained by choosing the most frequent rating across a total of 29 submitted summaries, for each episode and each question.
Our systems ``\textsf{\scriptsize{UCF\_NLP}}$\star$'' achieve the most gains compared to episode descriptions ``\textsf{\scriptsize{DESC}}'' on questions Q1 (people names), Q3 (main topics), Q5 (title context) and Q7 (good English).
}
\label{tab:human_eval_more}
\end{minipage}
\end{table*}

\begin{table}
\setlength{\tabcolsep}{5pt}
\renewcommand{\arraystretch}{1.15}
\centering
\begin{scriptsize}

\begin{minipage}{\textwidth}
\textsf{
\begin{tabular}{|l|rrr|c|rrr|}
\hline
\rowcolor{gray!15}
& \multicolumn{3}{c|}{\textcolor{blue}{\textbf{Majority Wins} (\%)}} & \textbf{Equal} & \multicolumn{3}{c|}{\textcolor{red}{\textbf{System Wins} (\%)}}\\
\rowcolor{gray!15}
System & -3 & -2 & -1 & 0 & +1 & +2 & +3\\
\hline
\hline
DESC & 0.0 & 2.2 & 8.4 & 48.0 & 22.9 & 12.8 & 7.3\\
FILT & 0.0 & 2.2 & 8.4 & 44.7 & 25.1 & 10.1 & 9.5\\
\textcolor{red}{UCF\_NLP1} & 0.6 & 1.1 & 5.0 & 43.6 & 22.9 & 19.0 & 7.8\\
\textcolor{red}{UCF\_NLP2} & 0 & 1.1 & 2.2 & 43.6 & 26.3 & 16.2 & 10.6\\
\hline
\end{tabular}
}
\end{minipage}
\vskip3\baselineskip
\begin{minipage}{\textwidth}
\textsf{
\begin{tabular}{|l|rrr|c|rrr|}
\hline
\rowcolor{gray!15}
& \multicolumn{3}{c|}{\textcolor{blue}{\textbf{DESC Wins} (\%)\phantom{\textbf{xx}}}} & \textbf{Equal} & \multicolumn{3}{c|}{\textcolor{red}{\textbf{System Wins} (\%)\phantom{\textbf{x}}}}\\
\rowcolor{gray!15}
System & -3 & -2 & -1 & 0 & +1 & +2 & +3\\
\hline
\hline
FILT & 1.1 & 1.1 & 14.0 & 65.9 & 15.1 & 2.2 & 0.6\\
\textcolor{red}{UCF\_NLP1} & 0.6 & 6.7 & 17.9 & 39.7 & 23.5 & 9.5 & 2.2\\
\textcolor{red}{UCF\_NLP2} & 0.6 & 5.0 & 15.1 & 41.9 & 24.0 & 10.6 & 2.8\\
\hline
\end{tabular}
}
\end{minipage}

\end{scriptsize}
\caption{
Percentages of testing episodes (out of a total of 179) on which a system performs equal to or better/worse than the majority baseline (``Majority'', top), and episode descriptions (``\textsf{\scriptsize{DESC}}'', bottom). 
``$\pm$1/2/3'' represents the gap between the numerical ratings, measured in terms of summary quality. 
The majority rating was obtained by choosing the most frequent quality rating across a total of 29 submitted summaries, for each episode.
Our system ``\textsf{\scriptsize{UCF\_NLP2}}'' frequently outperforms or performs comparably to the majority baseline (96.65\%) and episode descriptions (79.33\%).
}
\label{tab:majority_baseline}
\end{table}

\begin{table}
\setlength{\tabcolsep}{8pt}
\renewcommand{\arraystretch}{1.2}
\centering
\begin{scriptsize}
\textsf{
\begin{tabular}{|llrrr|}
\hline
\rowcolor{gray!15}
\textcolor{red}{\textbf{Test-1027}} & System & P (\%) & R (\%) & F (\%)\\
\hline
\hline
\multirow{2}{*}{ROUGE-1} & UCF\_NLP1 & 37.14 & 30.48 & 29.62\\
& UCF\_NLP2 & 36.29 & 31.39 & \textbf{29.64}\\
\hline
\hline
\multirow{2}{*}{ROUGE-2} & UCF\_NLP1 & 15.82 & 12.43 & \textbf{12.42}\\
& UCF\_NLP2 & 14.89 & 12.52 & 11.96\\
\hline
\hline
\multirow{4}{*}{ROUGE-L} & UCF\_NLP1 & 26.63 & 22.23 & \textbf{21.40}\\
& UCF\_NLP1$^\ddag$ & 26.67 & 22.05 & 21.32\\
& UCF\_NLP2 & 25.53 & 22.59 & 20.99\\
& UCF\_NLP2$^\ddag$ & 25.53 & 22.42 & 20.91\\
\hline
\end{tabular}
}
\end{scriptsize}
\caption{
ROUGE scores on the full test set containing 1,027 episodes.
The episode descriptions are used as gold-standard summaries; they are called ``model'' summaries.
$\ddag$ represents ROUGE scores computed by NIST. 
The other scores are calculated by us. 
There is a minor discrepancy between these scores when given the same summaries ($\pm$0.18\% for ROUGE-L).
\textsf{\scriptsize{UCF\_NLP1}} marginally outperforms \textsf{\scriptsize{UCF\_NLP2}} according to ROUGE-2 and ROUGE-L.
}
\label{tab:rouge_1027}
\end{table}

\begin{table}
\setlength{\tabcolsep}{8pt}
\renewcommand{\arraystretch}{1.2}
\centering
\begin{scriptsize}
\textsf{
\begin{tabular}{|llrrr|}
\hline
\rowcolor{gray!15}
\textcolor{red}{\textbf{Test-179}} & System & P (\%) & R (\%) & F (\%)\\
\hline
\hline
\multirow{2}{*}{ROUGE-1} & UCF\_NLP1 & 40.20 & 31.98 & \textbf{31.70}\\
& UCF\_NLP2 & 38.05 & 32.67 & 31.18\\
\hline
\hline
\multirow{2}{*}{ROUGE-2} & UCF\_NLP1 & 18.52 & 13.72 & \textbf{14.03}\\
& UCF\_NLP2 & 16.23 & 13.53 & 12.95\\
\hline
\hline
\multirow{2}{*}{ROUGE-L} & UCF\_NLP1 & 29.18 & 23.33 & \textbf{23.04}\\
& UCF\_NLP2 & 27.06 & 23.29 & 22.03\\
\hline
\end{tabular}
}
\end{scriptsize}
\caption{
ROUGE scores on the test set of 179 episodes.
The scores are calculated by us. 
Episode descriptions are used as gold-standard summaries; they are called ``model'' summaries.
We observe that \textsf{\scriptsize{UCF\_NLP1}} outperforms \textsf{\scriptsize{UCF\_NLP2}} in terms of R-1, R-2 and R-L F-scores.
We note that ROUGE scores are not necessarily a measure of summary quality, as
\textsf{\scriptsize{UCF\_NLP2}} was rated higher according to human judgments.
}
\label{tab:rouge_179}
\end{table}

\subsection{Summary Postprocessing}
\label{sec:length}

We next describe our efforts at postprocessing the summaries generated by BART.
The \textsf{\textbf{\footnotesize{length\_penalty}}} of BART penalizes the log probability of a summary (a negative value) by its length, with an exponent $p$,
setting $p$=2.0 promotes the generation of longer summaries. 
We set \textsf{\textbf{\footnotesize{no\_repeat\_ngram\_size}}} to be 3, which stipulates that a trigram cannot occur more than once in the summary. 
After a grid search in the range of [35,42], we set the \textsf{\textbf{\footnotesize{min\_length}}} of a summary to be 35 subwords for \textsf{{\footnotesize{UCF\_NLP1}}} and 39 for \textsf{{\footnotesize{UCF\_NLP2}}}.
The \textsf{\textbf{\footnotesize{max\_length}}} of a summary is set to 250 subwords.
We use a beam size of $K$=4 for summary decoding.

BART may optimize well with the inductive bias, 
but it remains necessary to apply a series of heuristics to the output summaries to alleviate any overfitting that has led to the generation of template language and improve the summary presentability.
Among others, the heuristics include 
(a) removing the content after ``---'' (e.g., ``\emph{--- This episode is sponsored by},'' ``\emph{--- Send in a voice message}'');
(b) removing URLs;
(c) removing brackets and the content inside it;
(d) removing any trailing incomplete sentence if the summary is excessively long ($\ge$128 tokens);
(e) removing duplicate sentences that occur three times or more across different episodes.
We observe that a handful of sentences appeared in summaries of different episodes.
They are unrelated to the transcripts, but are generated possibly due to overfitting (e.g., \emph{The Oops podcast examines the mistakes that change the trajectory of people's lives: the bad decisions, the aftermath, the path to redemption and all things in between.})

\section{Results and Conclusion}
\label{sec:future}

There were 29 submitted runs for the podcast summarization task of TREC 2020.
Each run contains a set of summaries generated for the full test set with 1,027 episodes.
Among these, 179 episodes were selected by NIST evaluators in a random fashion to perform qualitative judgments on the summary quality.
An evaluator quickly skimmed the episode, and made judgments for each summary for that episode, in a random order. 
Intermixed in the submitted summaries were the creator description (``\textsf{\footnotesize{DESC}}'') for an episode, and a ``filtered'' summary from Spotify (``\textsf{\footnotesize{FILT}}'').
Our system runs are denoted by ``\textsf{\footnotesize{UCF\_NLP1}}'' and ``\textsf{\footnotesize{UCF\_NLP2}}''.
The latter used an additional extraction module to identify salient segments from the transcripts.

The evaluation criteria used by NIST evaluators are shown in Table~\ref{tab:eval_criteria}.
The summary quality rating is a number from 0-3, corresponding to four levels: Bad/Fair/Good/Excellent.
Additionally, there were eight yes-or-no questions asked about the summary (Table~\ref{tab:eval_questions}).
They were designed to evaluate the various aspects of summaries.
An ideal summary will receive a ``yes'' for all questions but Q6.

We present our results in Tables~\ref{tab:human_eval}--\ref{tab:rouge_179}.
Our system ``\textsf{\footnotesize{UCF\_NLP2}}'' has achieved a quality rating of 1.559. 
This is an absolute increase of 0.268 (+21\%) over episode descriptions.
We raise awareness of possible inconsistencies between ROUGE and human judgments of summary quality.
The inconsistencies could stem from the deficiencies of ROUGE in capturing semantic meanings, or it could be due to episode descriptions are not the most appropriate reference summaries. 
We find content selection to remain important for podcast summarization.
A summary containing only partial information about an episode may hamper the reader's understanding.
Further, we caution that it is challenging to generate abstractive summaries that are fully accurate to the content of the podcast.
Not only are there transcription errors, but subtle change of meaning can happen in system abstracts.
Even humans may not spot some subtle errors without a careful comparison of system abstracts and transcripts.

\section*{Acknowledgments}
This research was supported in part by the National Science Foundation IIS-1909603.
We are grateful to Amazon for partially sponsoring the research and computation in this study through the Amazon AWS Machine Learning Research Award.

\bibliography{more,fei,anthology}
\bibliographystyle{acl_natbib}


\end{document}